\documentclass[letterpaper, 10 pt, conference]{ieeeconf}  

\IEEEoverridecommandlockouts                              

\usepackage{graphicx}
\usepackage{subfigure}
\usepackage{bm}
\usepackage{color}

\title{\LARGE \bf
A Systematic Evaluation of Different Indoor Localization Methods in Robotic Autonomous Luggage Trolley Collection at Airports 
}

\author{Zhirui Sun, Weinan Chen, Jiankun Wang, \emph{Senior Member, IEEE} and Max Q.-H. Meng, \emph{Fellow, IEEE} 
\thanks{The work in this article is supported by National Natural Science Foundation of China grant \#62103181, \emph{(Corresponding authors:Jiankun Wang, Max Q.-H. Meng).}}
\thanks{Zhirui Sun and Jiankun Wang are with Shenzhen Key Laboratory of Robotics Perception and Intelligence, and the Department of Electronic and Electrical Engineering, Southern University of Science and Technology, Shenzhen 518055, China, Jiaxing Research Institute, Southern University of Science and Technology, Jiaxing, China. {\tt\small e-mail: wangjk@sustech.edu.cn}}%
\thanks{Weinan Chen is with Guangdong University of Technology, Guangzhou, China.}%
\thanks{Max Q.-H. Meng is with Shenzhen Key Laboratory of Robotics Perception and Intelligence and the Department of Electronic and Electrical Engineering at Southern University of Science and Technology in Shenzhen, China. He is a Professor Emeritus in the Department of Electronic Engineering at The Chinese University of Hong Kong in Hong Kong and was a Professor in the Department of Electrical and Computer Engineering at the University of Alberta in Canada. {\tt\small e-mail: max.meng@ieee.org}}%
}

\begin{document}

\maketitle
\thispagestyle{empty}
\pagestyle{empty}

\begin{abstract}

This article addresses the localization problem in robotic autonomous luggage trolley collection at airports and provides a systematic evaluation of different methods to solve it. The robotic autonomous luggage trolley collection is a complex system that involves object detection, localization, motion planning and control, manipulation, etc. Among these components, effective localization is essential for the robot to employ subsequent motion planning and end-effector manipulation because it can provide a correct goal position. In this article, we survey four popular and representative localization methods to achieve object localization in the luggage collection process, including radio frequency identification (RFID), Keypoints, ultrawideband (UWB), and Reflectors. To test their performance, we construct a qualitative evaluation framework with Localization Accuracy, Mobile Power Supplies, Coverage Area, Cost, and Scalability. Besides, we conduct a series of quantitative experiments regarding Localization Accuracy and Success Rate on a real-world robotic autonomous luggage trolley collection system. We further analyze the performance of different localization methods based on experiment results, revealing that the Keypoints method is most suitable for indoor environments to achieve the luggage trolley collection.

\end{abstract}


\section{Introduction}
Currently, robots are widely use in our daily life, helping people handle various tasks encountered in life, such as sweeping, driving, and education. At present, the luggage trolleys at most airports need to be collected by human labor. Airports have to employ large numbers of people to collect and return the luggage trolleys to designated places for passengers to use. It is very labor-intensive and time-consuming. Therefore, using robots to collect luggage trolleys can significantly reduce the number of people involved in this work, improve the efficiency of the luggage trolley collection, and reduce airport management costs. However, the robotic autonomous luggage trolley collection is a complex system that integrates object detection, localization, motion planning, manipulation, etc. 
\begin{figure}[t]
    \centering
    \includegraphics[width=1\columnwidth]{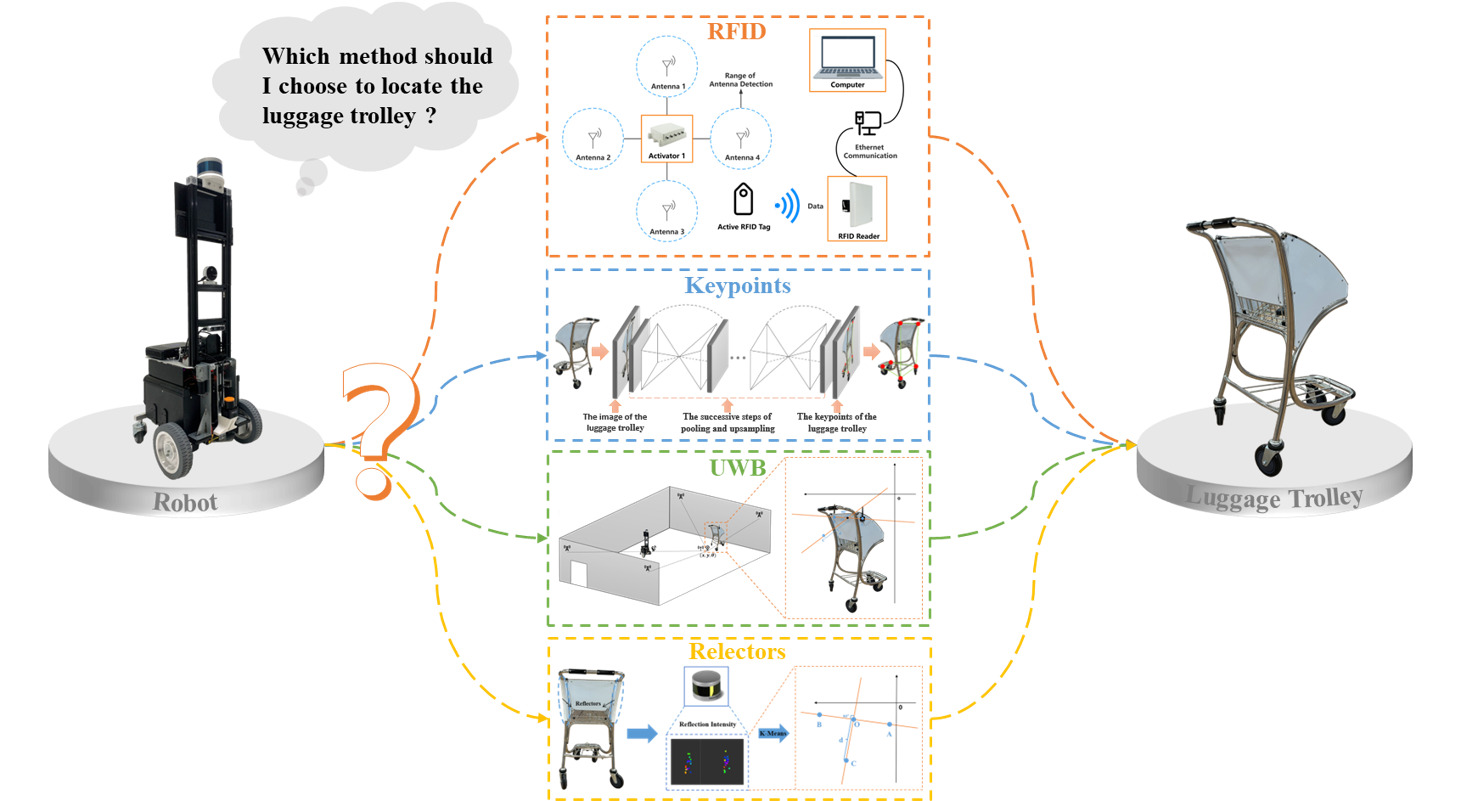}
    \caption{Different localization methods in robotic autonomous luggage trolley collection.}
    \label{method}
\end{figure}
In the luggage trolley collection, the robot needs to accurately reach the position behind the luggage trolley to achieve the necessary manipulation. Thus, an effective real-time localization method for the luggage trolley is crucial for robots to complete the luggage trolley collection at airports. At present, there are many localization methods, and each has its own characteristics and suitable scenarios.

The global positioning system (GPS) basically meets the need of localization for users in outdoor scenarios. Since the signal is attenuated and interrupted by the roofs and walls of the buildings, GPS is inefficient for indoor location-based services \cite{GPS}. However, 80\% of a person's life is spent indoors \cite{80}. Meanwhile, more and more skyscrapers, shopping malls, transportation hubs, stadiums, and medical centers are all over the city. A large number of localization requirements occur in the interior space of these modern buildings, such as personnel location, service robot, and indoor navigation. The value of indoor localization is becoming more and more evident. Academia and industry are gradually devoting a lot of effort and resources to this area. According to the type of sensors used, indoor localization methods can be roughly divided into three types:
\begin{itemize}
    \item Wireless sensor network localization: The sensors used in this method are RFID \cite{intro-rfid}, UWB \cite{intro-uwb}, etc. Generally speaking, the distance information between the sensors can be obtained through the ranging algorithm, such as TOA \cite{toa}, TDOA \cite{tdoa}, RSSI \cite{rssi}, etc. Then the geometric relationships are established to calculate the object's position. 
    \item Visual localization: The sensors used in this method are monocular, binocular, and RGB-D cameras. This method achieves localization by estimating the camera's pose from the captured image information. Visual simultaneous localization and mapping (VSLAM) is the famous research of visual localization.
    \item Laser localization: Herein, the sensor usually is 2D or 3D lidar. One of the laser localization methods is to construct a map with lidar in advance, and then the location information is calculated by matching the lidar data with the map features. In addition, lidar can also obtain information based on optical properties. Thus, localization can be achieved based on the apparent difference in the reflection intensity of reflectors.
\end{itemize}

The airport is a kind of indoor environment, and the localization of the luggage trolley requires an effective indoor localization method. According to the above sensor classification, this article systematically evaluates the performance of four popular and representative indoor localization methods, including RFID, UWB, Keypoints, and Reflectors in robotic autonomous luggage trolley collection at airports, as shown in Fig. \ref{method}.

The contributions of our study are two-fold:
\begin{itemize}
    \item We establish a systematic evaluation framework, including quantitative and qualitative metrics, to evaluate the performance of four typical indoor localization methods.
    \item The localization effect of these four methods is verified in the specific scenario of robotic autonomous luggage trolley collection. The verified results provide a valuable reference for the indoor localization of robots in other scenarios.
\end{itemize}

The rest of this article is organized as follows. Section II presents a review of related work. Section III details four indoor localization methods, including RFID, Keypoints, UWB, and Reflectors. The experiment setup, results, and discussion are explained in Section IV. Section V concludes this work and addresses the future direction. 
\section{Related Work}
Our work mainly focuses on effective indoor localization in robotic autonomous luggage trolley collection. To further clarify the motivation of this article, this section summarizes the research in this area and discusses the shortcomings of the existing research.

\subsection{Indoor Localization}
In \cite{challenge}, the authors compare different indoor localization systems and present some challenges faced by localization systems. Liu et al. \cite{liu} review some wireless indoor localization methods and discuss different performance measurement criteria and some trade-offs among them. In \cite{wireless-sensor}, localization methods for mobile wireless sensor networks are reviewed. This survey mainly focuses on indoor and outdoor wireless sensor networks, provides a classification for mobile wireless sensors and localization, and gives some related practical applications of mobile sensors. Davidson and Piché \cite{smartphones} mainly review related indoor localization methods on smartphones, including localization based on Wi-Fi, Bluetooth, magnetic field fingerprinting, etc. A detailed overview of localization systems for emergency responders is provided in \cite{emergency}. This review mainly discusses different indoor localization methods and their strengths and weaknesses in emergency response systems. Faheem et al. \cite{2019-survey} present a detailed review of different indoor localization techniques, technologies, and systems. Besides, they propose an evaluation framework to evaluate different indoor localization systems. The localization system technologies, indoor localization techniques, localization detection techniques, and wireless technologies are introduced in \cite{2021-review}.

However, most of the existing literature is a review of indoor localization methods, which focus on comparative analysis of principles without building an evaluation framework for systematic evaluation. Although a few kinds of literature have evaluated indoor localization methods, but there is no factual experimental verification. Therefore, in this work, we establish a systematic evaluation framework and design the real-world experiment to verify the performance of four popular indoor localization methods.

\subsection{Robotic Autonomous Luggage Trolley Collection}

The first solution of robotic autonomous luggage trolley collection is proposed in \cite{trolley-first}. They present a luggage trolley pose estimation method based on point cloud matching. Wang et al. \cite{real-time-plan} introduce a novel real-time path planning algorithm in robotic autonomous luggage trolley collection. They verify the effectiveness of the proposed algorithm in simulation experiments. A novel mobile manipulation system with applications in robotic autonomous luggage trolley collection is presented by Xiao et al. \cite{keypoints}. In localization, they use a keypoint detection net and the Efficient Perspective-n-Point (EPnP) algorithm \cite{epnp} to get a 6D pose of the luggage trolley. Eventually, combined with the robot's pose, the state of the luggage trolley can be obtained.

The robotic autonomous luggage trolley collection is a complex system, and it is challenging to address all issues in one study. The existing literature mainly focuses on the path planning algorithm and the design of the entire system. Although the indoor localization method based on vision is adopted to the luggage trolley, there is no complete comparative analysis, and the effectiveness of the luggage trolley localization has not yet been evaluated. 

In response to the above problems, we select four representative methods, including RFID, UWB, Keypoints and Reflectors for evaluation and analysis. Based on the robotic autonomous luggage trolley collection, we design experiments on the real robot system to evaluate the performance of these four methods.

\section{Four Indoor Localization Methods}

This section presents the main features of the four indoor localization methods we have selected for evaluation: RFID, UWB, Keypoints, and Reflectors. These four methods represent three different sensors used in indoor localization. The RFID and UWB methods depend on the wireless sensor. While, the Keypoints and Reflectors methods are based on visual and laser sensors, respectively. 

\subsection{RFID Localization Method}
\begin{figure}[htb]
    \centering
    \includegraphics[width=1\columnwidth]{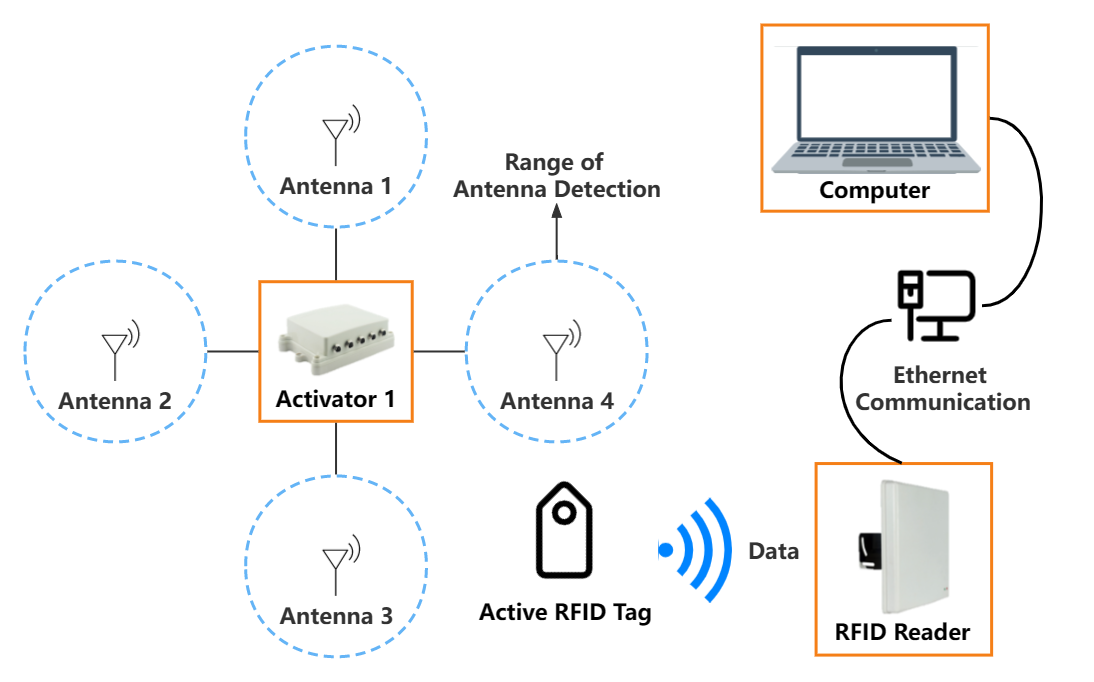}
    \caption{The schematic diagram of the RFID method.}
    \label{rfid}
\end{figure}

RFID has the characteristics of non-contact communication, high data rate, and low cost \cite{rfid-feature}, which make it be a good candidate for indoor localization. The principle of RFID is using radio frequency to conduct data communication between the RFID reader and the tag to achieve the purpose of identifying and tracking objects. RFID tags are divided into active and passive tags \cite{rfid-tag}, and we use active tags in our experiment. With the RFID method, the location information of the receiving signal needs to be known in advance. When locating the luggage trolley, since the position of the luggage trolley is not fixed, the location information of the tag cannot be set in advance. Thus the real-time position of the luggage trolley cannot be obtained simply by collecting the tag information through the RFID reader. We set the position and the unique number of the antenna, and the localization of the luggage trolley can be obtained by receiving the unique number information of the antenna near the luggage trolley. In this way, the RFID system consists of an activator with four antennas, an RFID reader, an active tag, and a computer, as shown in Fig. \ref{rfid}.

One activator has four antennas, each with a corresponding position and a unique number. When the active tag is close to the antenna, it will be activated and carry the unique number of this antenna. When the active tag sends out a signal, it will also send this unique number. The system judges the current position of the tag through the strength and the antenna's unique number of the detected tag signal. The localization accuracy of this RFID system mainly depends on the range of antenna detection.

\subsection{Keypoints Localization Method}
The 6D pose estimation of objects is the key to many applications in the real world, especially for robot manipulation and grasping \cite{manipulation, grasping}. 

\begin{figure}[htb]
\centering
    \includegraphics[width=1\columnwidth]{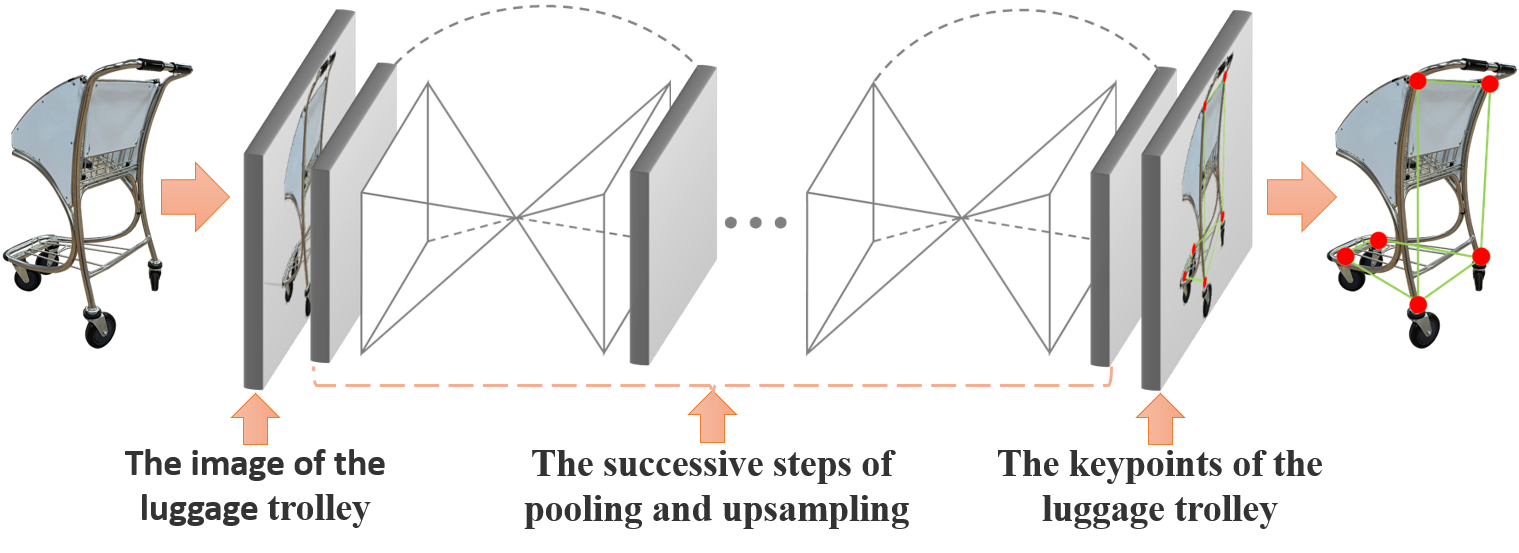}
    \hspace{0.5cm}
\caption{The schematic diagram of the Keypoints method.}
\label{keypoints-diagram}
\end{figure}

As shown in Fig. \ref{keypoints-diagram}, the six red 2D keypoints ($p_k = [x_k, y_k]^T$, $k=0, 1,..., 5$) can be estimated by the stacked hourglass network structure. Then, we can solve the Perspective-n-Point (PnP) problem through the EPnP algorithm to obtain the 6D pose of the luggage trolley. The 6D pose consists of 3D rotation $R$ and 3D translation $T$ from the luggage trolley's frame to the camera's frame. Finally, combined with the localization of the robot, the state of the luggage trolley $S_{trolley} = [x, y, \theta]^T$ can be calculated.

\subsection{UWB Localization Method}
\begin{figure}[htb]
    \centering
    \includegraphics[width=1\columnwidth]{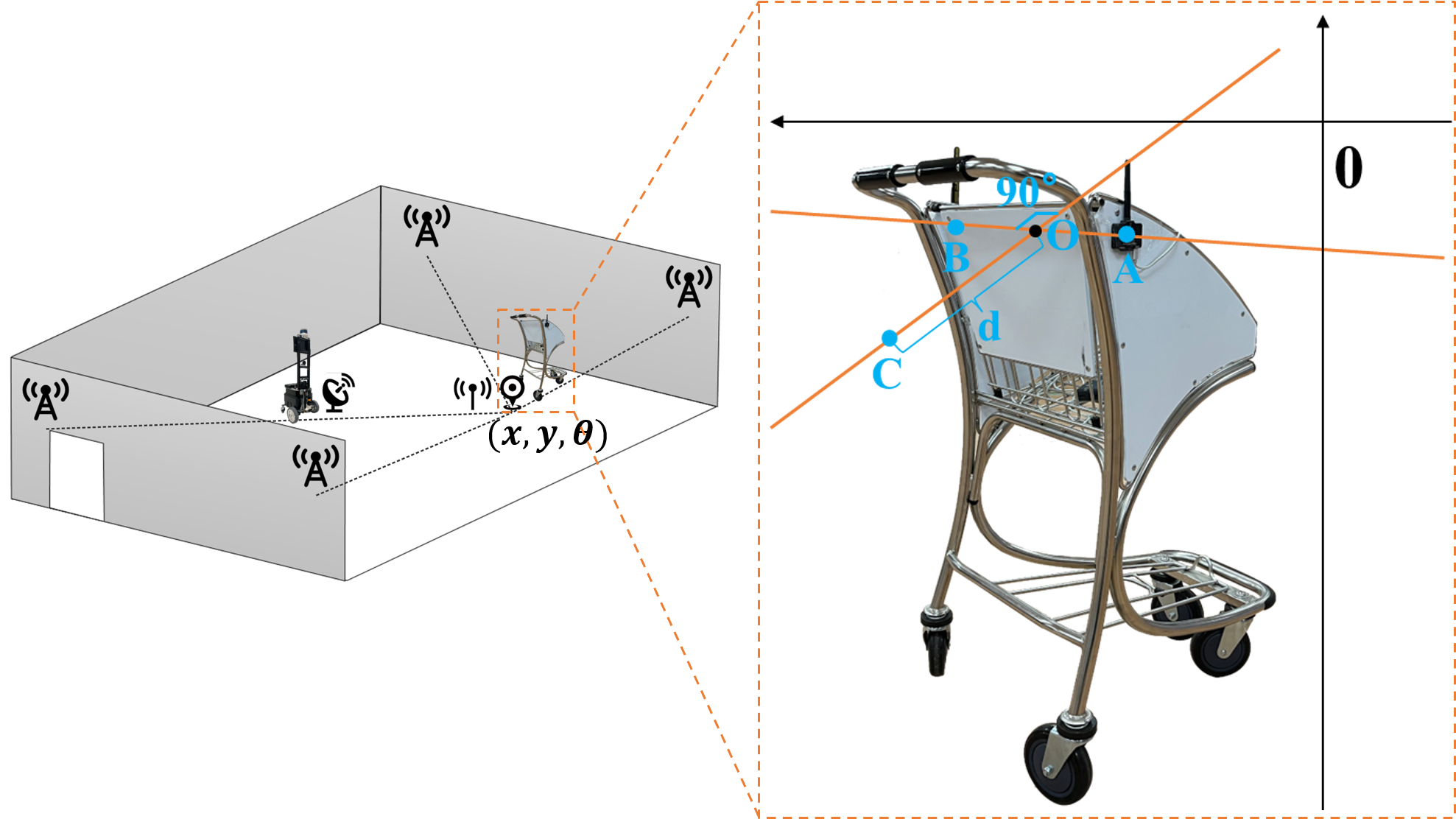}
    \caption{The schematic diagram of the UWB method.}
    \label{uwb}
\end{figure}

UWB is a radio frequency technology that has received extensive attention for indoor precise localization in recent years. LinkTrack is currently an advanced indoor localization system based on UWB technology. There has been a lot of localization research based on the LinkTrack system. For example, Cao et al. \cite{mobile-robot} present a solution to estimate the relative localization of mobile robots. A multi-agent 2D relative pose localization approach based on the LinkTrack system is proposed by Fishberg and P. How \cite{multi-agent}. Similarly, based on the LinkTrack system, we realize the indoor localization of the luggage trolley. The LinkTrack system is shown in Fig. \ref{uwb}. The four base stations at the room corners can locate the tag attached on the luggage trolley. Then the tag sends the localization information to the robot to help obtain the location of the luggage trolley.

The single-label UWB system can only obtain the $x$ and $y$ coordinates of the object. To get angle information, double labels are used in the UWB system.

Two labels are fixed on both sides of the luggage trolley, and they are set as points $A$ and $B$, respectively. The coordinates of points $A$ and $B$ are
\begin{equation}
\vec{A}=\left[\begin{array}{l}
x_A \\
y_A
\end{array}\right], \quad \vec{B}=\left[\begin{array}{l}
x_B \\
y_B
\end{array}\right].
\end{equation}

$\vec{O}$ is the midpoint between $\vec{A}$ and $\vec{B}$, and the coordinates of $\vec{O}$ are
\begin{equation}
\vec{O}=\frac{\vec{A}+\vec{B}}{2}=\left[\begin{array}{c}
\frac{x_A+x_B}{2} \\
\frac{y_A+y_B}{2}
\end{array}\right].
\end{equation}

$\vec{U}$ is the unit vector from $\vec{A}$ to $\vec{B}$, and the coordinates of $\vec{U}$ are
\begin{equation}
\vec{U}=\frac{\vec{B}-\vec{A}}{\|\vec{B}-\vec{A}\|}=\left[\begin{array}{c}
\frac{x_B-x_A}{\sqrt{\left(x_B-x_A\right)^2+\left(y_B-y_A\right)^2}} \\
\frac{y_B-y_A}{\sqrt{\left(x_B-x_A\right)^2+\left(y_B-y_A\right)^2}}
\end{array}\right]=\left[\begin{array}{l}
x_U \\
y_U
\end{array}\right].
\end{equation}

The unit vector $\vec{V}$ is obtained by rotating $\vec{U}$ 90 degrees counterclockwise, and the coordinates of $\vec{V}$ are
\begin{equation}
\vec{V}=\left[\begin{array}{c}
y_U \\
-x_U
\end{array}\right].
\end{equation}

Therefore, if $\vec{C}$ is at a distance of $d$ in the unit direction of $\vec{O}$, the coordinates of $\vec{C}$ are
\begin{equation}
\vec{C}=\vec{O}+d \vec{V}=\left[\begin{array}{l}
\frac{x_A+x_B}{2}+d y_U \\
\frac{y_A+y_B}{2}-d x_U
\end{array}\right].
\end{equation}
   
\subsection{Reflectors Localization Method}
The reflector is a highly reflective material that can reflect the incoming light toward the light source. The more light the reflector reflects, the greater the reflection intensity of the reflector will be. Based on reflection intensity, we can use lidar to distinguish the reflectors from ordinary objects \cite{reflector}. Because the installation and maintenance of reflectors are simple and convenient, they can be widely used in various scenarios. Therefore, identifying reflectors by lidar is also an effective indoor localization method.

\begin{figure}[htb]
    \centering
    \includegraphics[width=1\columnwidth]{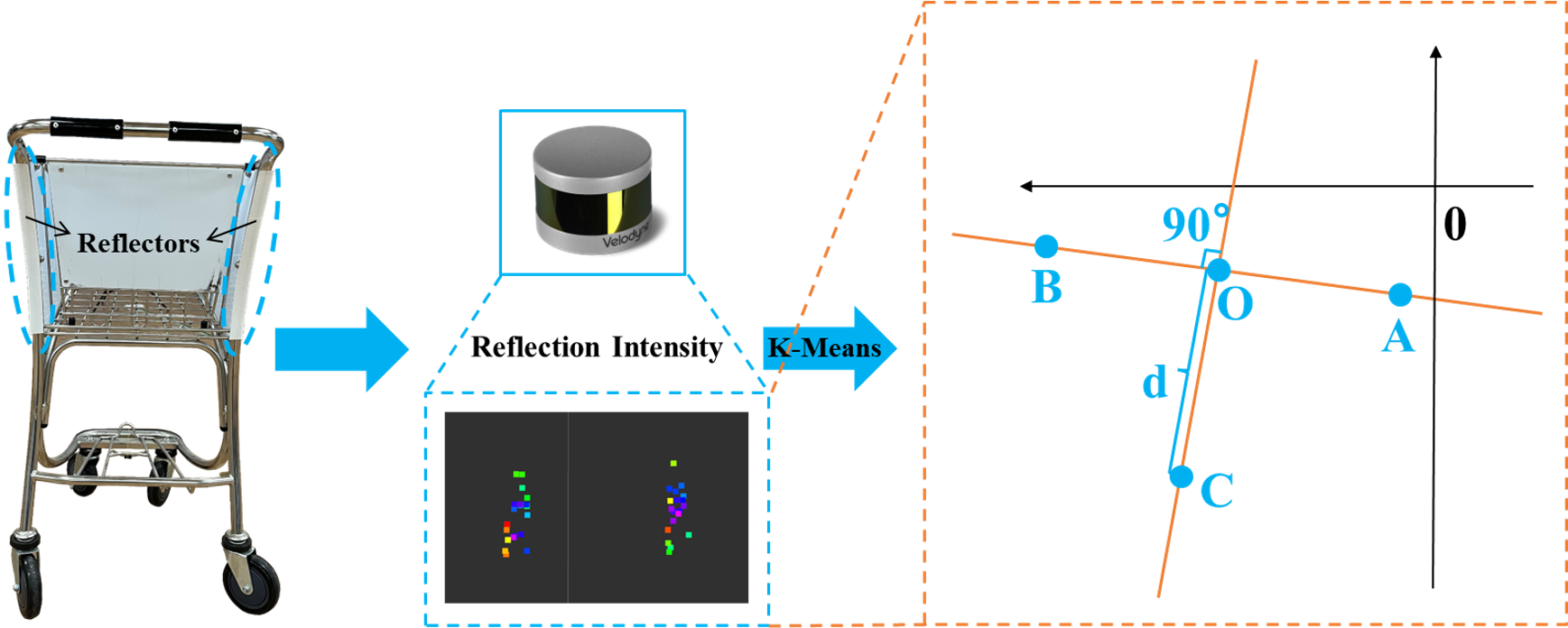}
    \caption{The schematic diagram of the Reflectors method.}
    \label{reflector}
\end{figure}

The point cloud of the reflector can be obtained by setting a reflection intensity threshold to filter. As shown in Fig. \ref{reflector}, we filter the point cloud acquired by the lidar to get the point clouds of two reflectors attached to the pole behind the luggage trolley. Then, the obtained two-part point clouds are clustered by the K-Means method \cite{kmeans}, and the center points of these two-part clouds are recorded as points $A$ and $B$. Next, the method of obtaining the coordinates of point $C$ is the same as the method of UWB above.

\section{Experiments and Results}
In this section, we evaluate these four indoor localization methods through qualitative and quantitative experiments. The localization error of the RFID method is in meters, and the angle information of the luggage trolley cannot be obtained. Only using RFID for localization cannot complete the task of collecting luggage trolleys. Therefore, we only do qualitative analysis for the RFID method.

\subsection{Experiment Setup}
In qualitative experiments, we set Localization Accuracy, Mobile Power Supplies, Coverage Area, Cost, and Scalability to evaluate these four indoor localization methods. In quantitative experiments, we test the performance of the UWB, Keypoints and Reflectors methods in terms of Localization Accuracy and Success Rate.

\begin{figure}[htb]
    \centering
    \includegraphics[width=1\columnwidth]{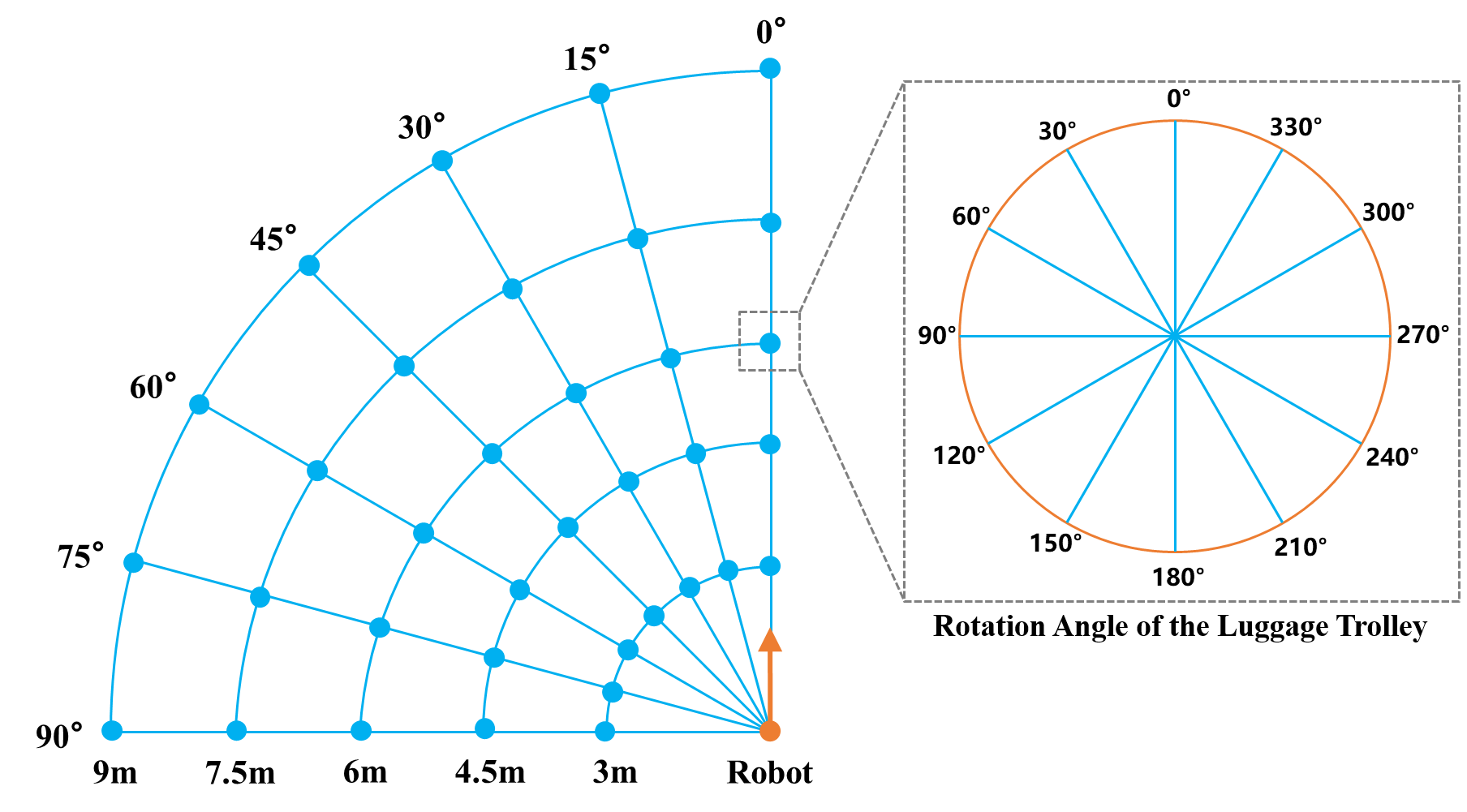}
    \caption{Different initial states of the luggage trolleys. The orange point is the pose of the robot. The blue point represents the different initial poses of the luggage trolley. Each blue point is similar to a polar coordinate, with different distances and rotation angles, as shown in the fan shape. The luggage trolley has 12 rotation angles at each blue point, as shown in the square.}
    \label{trolley_rotation}
\end{figure}

The sensing range of the Keypoints, UWB, and Reflectors methods are different in distance and angle in localization. When comparing the localization accuracy of the Keypoints, UWB, and Reflectors methods, we choose the point where all three methods can complete the localization. 
In the experiment of statistical success rate, we fix the robot's position and change the distance and angle of the luggage trolley, as shown in Fig. \ref{trolley_rotation}. In terms of the angle, we divide the quarter circle at intervals of 15°, and in terms of distance, we divide the diameter of the quarter circle at intervals of 1.5m from 3m to 9m. Besides, on the polar coordinates obtained by each combination of angle and distance, the luggage trolley rotates 360° at intervals of 30°. According to the symmetry, the success rate of the entire circular space can be represented by this quarter circle.

\subsection{Experiment Results}
\subsubsection{Qualitative Experiment Results} We evaluate these four methods on several qualitative metrics of Localization Accuracy, Mobile Power Supplies, Coverage Area, Cost and, Scalability. The results are shown in Table \ref{tab:qualitative}. 
\begin{itemize}
    \item Localization Accuracy: One of the most critical characteristics of indoor localization methods is the localization accuracy. Especially in robotic grasping operations, a good indoor localization method should be able to locate the object within a range of 10cm (known as $microlocation$).
    \item Mobile Power Supplies: In the indoor localization method, the part that consumes a lot of power should be moved to the server or an entity that can obtain an uninterruptible power supply. Due to the limitations of communication, some indoor localization methods still need to be equipped with mobile power.
    \item Coverage Area: We hope that the indoor localization method can achieve effective localization in large indoor spaces such as hospitals, shopping malls, airports, etc. High coverage area can reduce the equipment required and thus reduce the cost of money in localization.
    \item Cost: The cost of money in the indoor localization method should not be high. In particular, an ideal method should have no additional costs, including extra equipment and installation costs. The accuracy of localization can be improved by adding proprietary hardware devices, but it also leads to additional costs.
    \item Scalability: A localization system should be scalable, meaning that the accuracy and real-time localization are not affected by the increase in localization objects. At the same time, it is also necessary to consider the cost caused by the rise in localization objects.
\end{itemize}

Regarding Localization Accuracy, the RFID method is about 1-4$m$ and the other three methods are all within 10$cm$, as shown in Table \ref{tab:qualitative}. The more accurate comparison among the Keypoints, UWB, and Reflectors methods is explained in subsequent quantitative experiments. 

Mobile Power Supplies means additional mobile power required in luggage trolley collection. The Keypoints method requires no extra mobile power because it requires no additional equipment other than the camera installed on the robot. Although the RFID method uses an active tag, the tag has its own battery, which lasts for a long time. The UWB method needs to communicate between multiple tags, and each tag needs to be equipped with a mobile power supply. The comparison of the four methods related to mobile power supplies is shown in Table \ref{tab:qualitative}. 

\begin{figure}[htb]
    \centering
    \includegraphics[width=1\columnwidth]{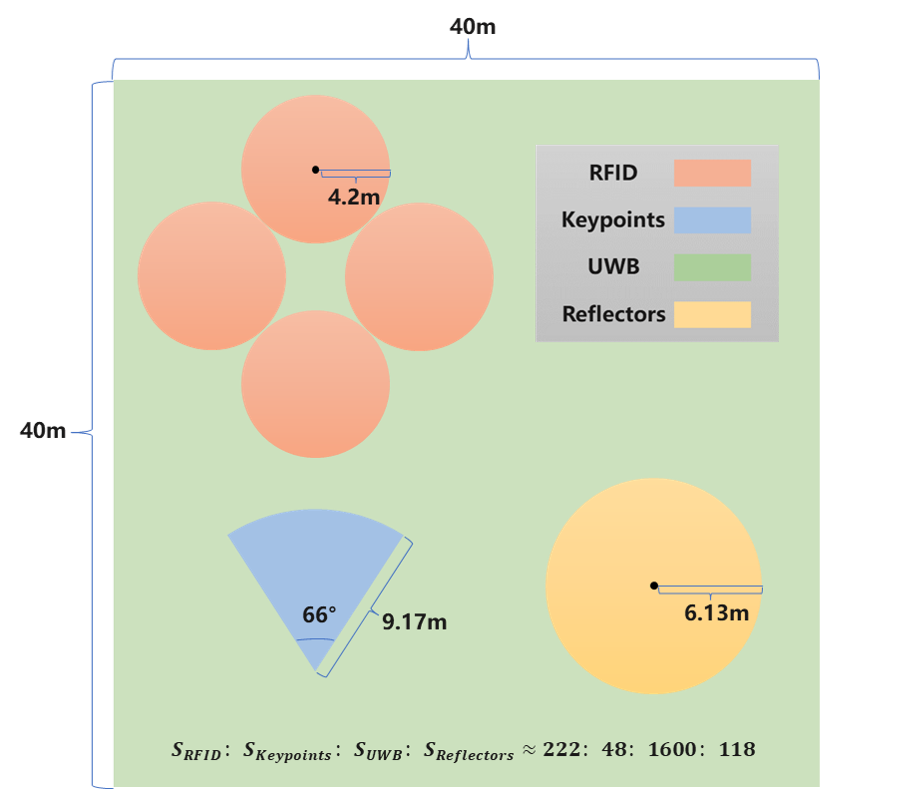}
    \caption{The coverage area of the four methods.}
    \label{area}
\end{figure}

The activator of the RFID method has four antennas, each antenna covers about 55.39$m^2$, and the total coverage area is about 222$m^2$. The coverage area of the Keypoints method is about 48$m^2$. In theory, the coverage area of the UWB method is about 40$m$ $\times$ 40$m$, which is 1600$m^2$. The coverage area of the Reflectors method is about 118$m^2$. The area comparison of the four methods is shown in Fig. \ref{area}.

We count the cost of the equipment. The cost of the RFID and UWB methods is about US\$500 (one activator with four antennas, one RFID reader, and one active tag) and US\$700 (seven tags, four supports, and four mobile power supplies), respectively. The robot needs to achieve global localization through lidar, so the Reflectors method does not need additional lidar. The cost of the Reflectors method is about US\$3 (two reflectors), and the cost of the Keypoints method is about US\$400 (one camera). RFID and UWB methods need to build additional systems, and the Reflectors method needs to stick reflectors. The Keypoints method needs to install the camera once. The installation of the Reflectors and Keypoints methods is simple and inexpensive. Installing the RFID and UWB methods is troublesome and requires much cost. Therefore, it is not difficult to find that the total cost of the RFID and UWB methods is high, and the total cost of the Reflectors and Keypoints methods is low and middle, respectively, as shown in Table \ref{tab:qualitative}.

When the number of localization objects increases, both RFID and UWB methods need to increase equipment to ensure the accuracy and effectiveness of localization. For example, each additional luggage trolley needs to be equipped with a corresponding tag. Besides, if the distribution space of the luggage trolley increases, corresponding reference points and anchors need to be added. However, the Keypoints method does not need to add any equipment as long as the keypoints of the luggage trolley remains unchanged, and the Reflectors method needs to paste the corresponding reflectors on the luggage trolley. Therefore, the scalability of the RFID and UWB methods is low, and the scalability of the Keypoints and Reflectors method is high and middle, respectively, as shown in Table \ref{tab:qualitative}.

\begin{table}[htb]
\centering
\caption{The results of four methods in qualitative metrics.}
\label{tab:qualitative}
\resizebox{0.46\textwidth}{!}{%
\begin{tabular}{ccccc}
\hline
Methods                 & RFID & Keypoints & UWB & Reflectors \\ \hline
Localization   Accuracy & 1-4m          & \textbf{1-10cm}             & \textbf{1-10cm}       & \textbf{1-10cm}             \\ 
Mobile Power   Supplies & \textbf{No}            & \textbf{No}                 & Yes          & \textbf{No}                 \\ 
Coverage Area           & $222m^2$           & $48m^2$                 & \bm{$1600m^2$}         & $118m^2$                \\ 
Cost                    & High          & Middle                  & High         & \textbf{Low}                \\ 
Scalability             & Low           & \textbf{High}               & Low          & Middle               \\ \hline
\end{tabular}
}
\end{table}

\subsubsection{Quantitative Experiment Results} We calculate the Mean Absolute Error (MAE) and Root Mean Square Error (RMSE) between the coordinates observed by the UWB, Keypoints, and Reflectors methods and the coordinates of ground truth. The formulas of MAE and RMSE are as follows:

\begin{equation}
MAE= \frac{1}{n} \sum_{i=1}^n(\left|x_i-\hat{x}_i\right| + \left|y_i-\hat{y}_i\right|)
\end{equation}

\begin{equation}
RMSE=\sqrt{\frac{1}{n} \sum_{i=1}^n[\left(x_i-\hat{x}_i\right)^2 + \left(y_i-\hat{y}_i\right)^2]}
\end{equation}

The results are shown in Table \ref{tab:precise-x}. Through the calculation results of MAE and RMSE, we can find that the $xy$ coordinates obtained by the Reflectors method are the most accurate, followed by the Keypoints method, and finally, by the UWB method. We analyze the reason that when the lidar of the Reflectors method detects the reflectors, there is not much error introduced. The Keypoints method can produce solution errors when estimating the camera pose. In the UWB method, errors will occur due to white noise when the signal is transmitted, and it is not easy to avoid.
\begin{table}[htb]\tiny
\centering
\caption{The MAE and RMSE results of three methods in xy-coordinate}
\label{tab:precise-x}
\resizebox{0.46\textwidth}{!}{%
\begin{tabular}{cccc}
\hline
Methods & Keypoints & UWB & Reflectors \\ \hline
MAE    & 0.1386             & 0.1813       & \textbf{0.1211}    \\ 
RMSE   & 0.1167             & 0.1622       & \textbf{0.1083}    \\ \hline
\end{tabular}%
}
\end{table}

\begin{figure}[htb]

	\centering
	\subfigure[The success rate results at 0°.]{
		\begin{minipage}{0.5\textwidth} 
                \includegraphics[width=\linewidth]{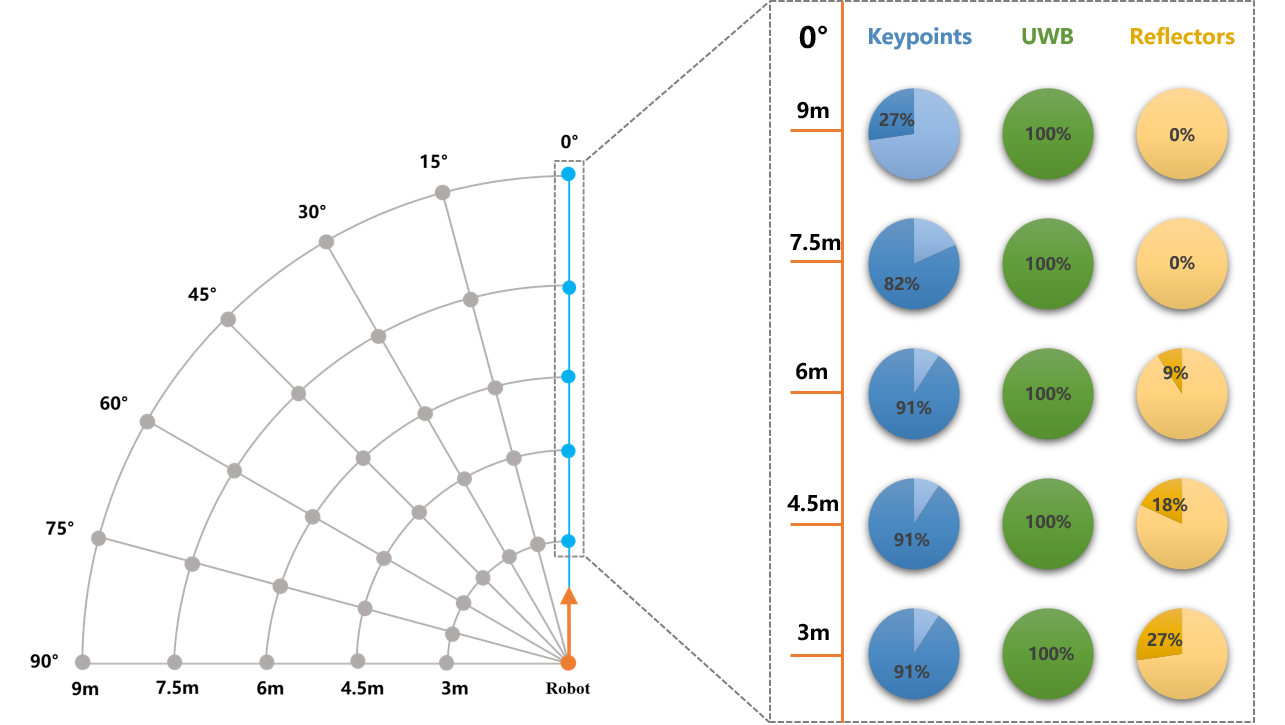} \\
			
		\end{minipage}
	}
	\subfigure[The success rate results at 0°, 30° and 45°.]{
		\begin{minipage}{0.5\textwidth}
			\includegraphics[width=\linewidth]{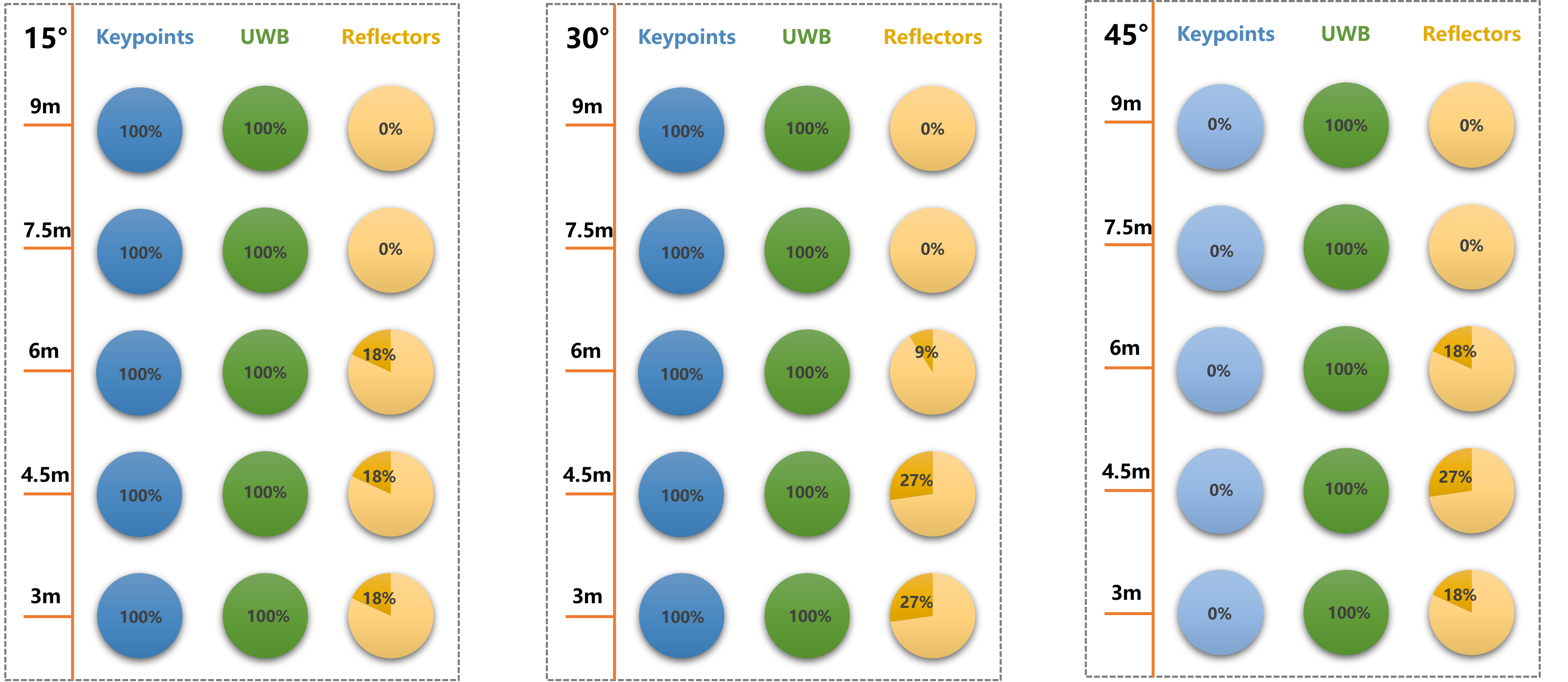} \\
			
		\end{minipage}
	}
 	\subfigure[The success rate results at 60°, 75° and 90°.]{
		\begin{minipage}{0.5\textwidth}
			\includegraphics[width=\linewidth]{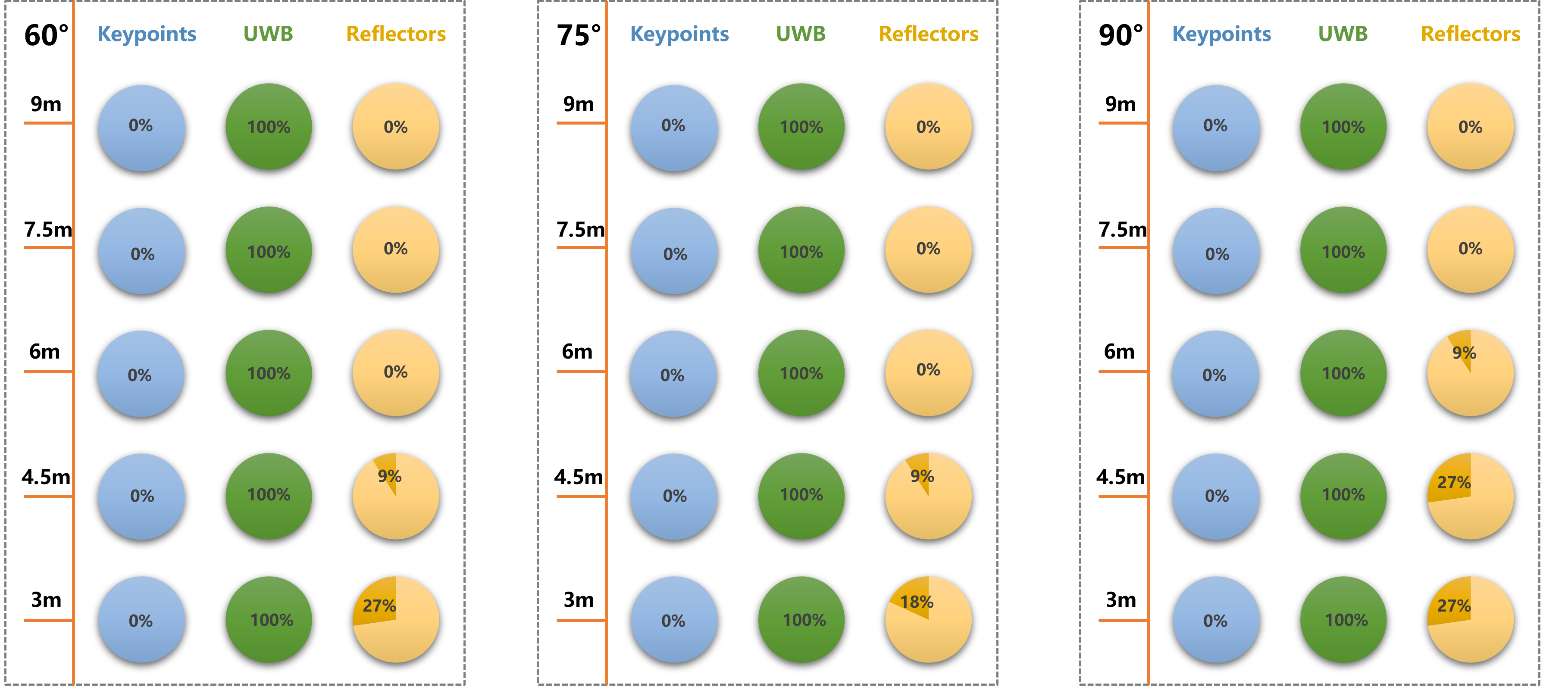} \\
			
		\end{minipage}
	}
\caption{Success rate results. The blue, green, and yellow circles represent the success rate results of the Keypoints, UWB, and Reflectors methods, respectively.} 
\label{rate}
\end{figure}
\begin{figure*}[htb]
\centering
    \includegraphics[width=2\columnwidth]{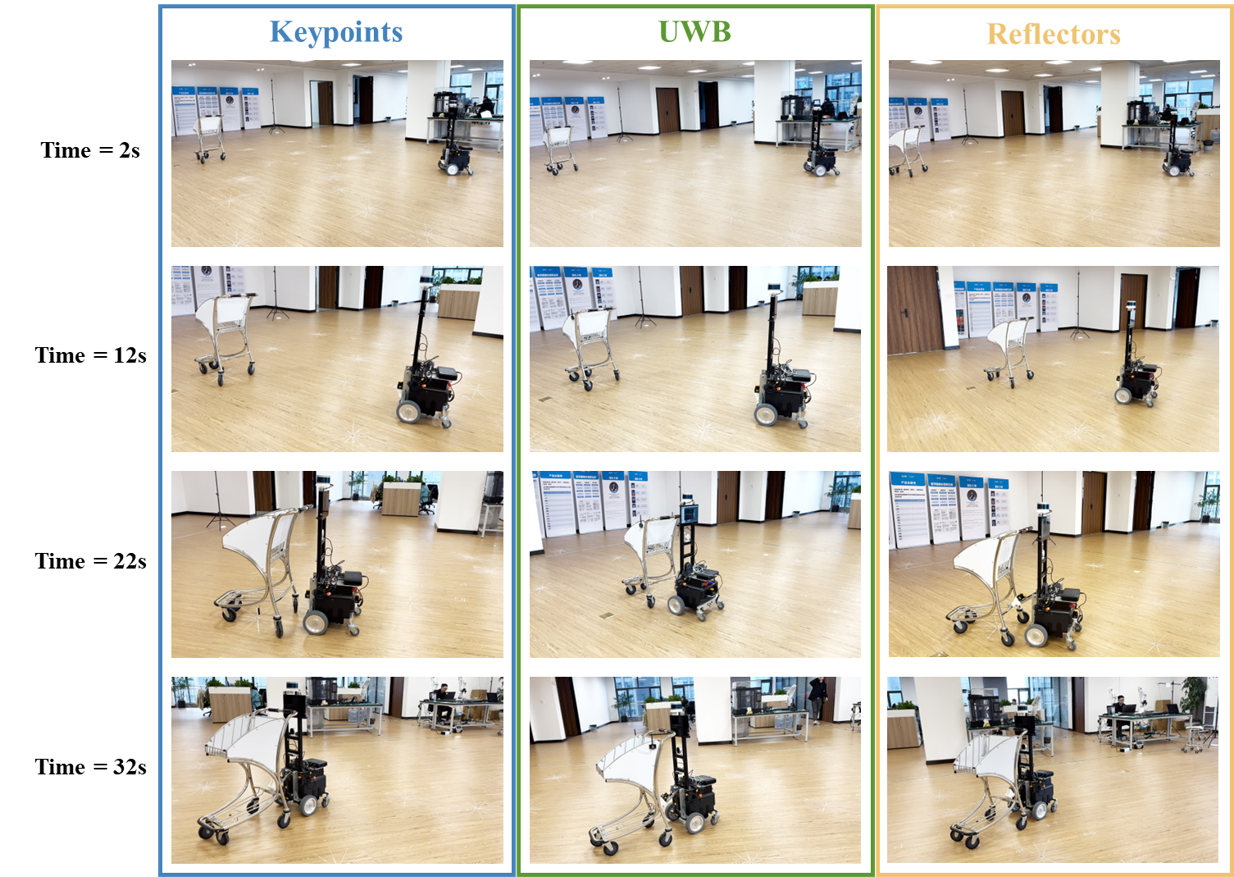}
    \hspace{0.5cm}
\caption{Screenshots of part of the real-world experiment process. We intercept pictures at four-time points, namely 2s, 12s, 22s, and 32s, which basically illustrate the whole collection process. The blue box, green box, and orange box represent the collection process of the Keypoints, UWB, and Reflectors methods, respectively.}
\label{real-world}
\end{figure*}

At each polar coordinate point, we calculate the success rate of each method to complete the collection task under different angles of the luggage trolley. At a certain rotation angle of the luggage trolley, we only consider two results: success or failure. Then the success rate of the luggage trolley collected by the robot at this polar coordinate point can be counted, and the results are shown in Fig. \ref{rate}. Part of the real-world experimental process is shown in Fig. \ref{real-world}.

From the results, we can find that the success rate of the UWB method is 100\% from 0° to 90°. This is because the coverage area of the UWB method is a space of 40$m$ $\times$ 40$m$, which is much larger than our experimental range. And the localization accuracy of the UWB method can meet the requirements of the luggage trolley collection. The success rate of Keypoints and Reflectors methods is affected by the distance between the robot and the luggage trolley and the angle of the luggage trolley. 
The success rate of the Keypoints method is 0\% from 45° to 90°. Because this range is beyond the camera's field of view, the keypoints of the luggage trolley cannot be obtained. In addition, even if the luggage trolley is within the camera's field of view, the keypoints information cannot be accepted due to the occlusion of the luggage trolley itself. For example, the success rate is 91\% at 0° because the occlusion of the luggage trolley itself affects the recognition of the keypoints at some angles. The occlusion effect will be exacerbated as the distance increases and the success rate decreases even lower. 
Because the reflectors are pasted on the back of the luggage trolley, the lidar can only obtain the information of the reflectors on the back of the luggage trolley, resulting in a low success rate of the Reflectors method. At the same time, it is evident that the recognition of the reflectors is affected by the distance, and the information of the reflectors' intensity cannot be obtained beyond a certain distance.

\subsection{Discussion}

These four methods have their characteristics and applicable scenarios. Although both RFID and UWB methods use the wireless sensor, the RFID method cannot meet the requirements in scenarios with high requirements of localization accuracy. If the accuracy requirements are not high, but the mobile power supply cannot be provided, the RFID method is more suitable than the UWB method.

The Keypoints, UWB, and Reflectors methods can meet the localization accuracy of the luggage trolley collection. These three methods also represent the use of three different sensors: the wireless sensor, the visual sensor, and the laser sensor. The Keypoints method does not require mobile power supplies, and the localization of the Keypoints method is relatively stable and accurate within the camera's field of view. Due to the extra hardware equipment, the UWB method can accurately and effectively realize object localization. Still, the most significant limitation of this method is that additional tags are required, and each tag requires a mobile power supply. The localization capability of the Reflectors method is the most accurate, but this method is greatly affected by the reflectors' pasting position and the sensor's recognition distance.

In robotic autonomous luggage trolley collection at airports, the localization accuracy of the luggage trolley is the first issue that needs to be considered, so the RFID method is unsuitable for this scenario. In addition, there are so many luggage trolleys at the airport, and adding mobile power suppliers to each luggage trolley is a huge cost, so the UWB method is not suitable. The environment of the airport is complex, and the placement of the luggage trolleys is messy. The localization method needs to be robust, which can adapt to the airport's complexity, and handle various uncertainties \cite{cleaning}. Besides, it is unrealistic to paste reflectors at all angles of the luggage trolley, so the Reflectors method is unsuitable. Therefore, among the several methods that can accomplish the luggage trolley collection, the Keypoints method is undoubtedly the most practical considering Localization Accuracy, Mobile Power Supplies, Coverage Area, Cost, Scalability, and Success Rate. Although the methods compared in this article are limited, each method is very representative, and they represent different sensor applications. In a single-sensor comparison, the vision-based method is best suitable for the robotic autonomous luggage trolley collection.

\section{Conclusions and Future Work}

In this article, we systematically evaluate the performance of four popular and representative indoor localization methods through qualitative and quantitative experiments. According to the analysis of experiment results, we find that the Keypoints method is the most suitable for the robotic autonomous luggage trolley collection. Under specific scenarios and tasks, the emphases and requirements for localization methods are different. As we have shown above, each method has its applicable scenarios. The analysis and experiment results of this article, especially the verification experiments in the specific scenario of robotic autonomous luggage trolley collection, can provide reference for the indoor localization of robots in other scenarios. 

In future work, we plan to implement these popular localization methods in real airports, such as Hong Kong International Airport and Shenzhen Baoan International Airport, to further test their performance and develop the most suitable localization method to achieve robotic autonomous luggage trolley collection. 

\bibliographystyle{IEEEtran} 

\bibliography{refs} 

\begin{thebibliography}{10}
\providecommand{\url}[1]{#1}
\csname url@samestyle\endcsname
\providecommand{\newblock}{\relax}
\providecommand{\bibinfo}[2]{#2}
\providecommand{\BIBentrySTDinterwordspacing}{\spaceskip=0pt\relax}
\providecommand{\BIBentryALTinterwordstretchfactor}{4}
\providecommand{\BIBentryALTinterwordspacing}{\spaceskip=\fontdimen2\font plus
\BIBentryALTinterwordstretchfactor\fontdimen3\font minus
  \fontdimen4\font\relax}
\providecommand{\BIBforeignlanguage}[2]{{%
\expandafter\ifx\csname l@#1\endcsname\relax
\typeout{** WARNING: IEEEtran.bst: No hyphenation pattern has been}%
\typeout{** loaded for the language `#1'. Using the pattern for}%
\typeout{** the default language instead.}%
\else
\language=\csname l@#1\endcsname
\fi
#2}}
\providecommand{\BIBdecl}{\relax}
\BIBdecl

\bibitem{GPS}
S.~Nirjon, J.~Liu, G.~DeJean, B.~Priyantha, Y.~Jin, and T.~Hart, ``Coin-gps:
  Indoor localization from direct gps receiving,'' in \emph{Proceedings of the
  12th annual international conference on Mobile systems, applications, and
  services}, 2014, pp. 301--314.

\bibitem{80}
N.~Singh, S.~Choe, and R.~Punmiya, ``Machine learning based indoor localization
  using wi-fi rssi fingerprints: An overview,'' \emph{IEEE Access}, vol.~9, pp.
  127\,150--127\,174, 2021.

\bibitem{intro-rfid}
F.~Tlili, N.~Hamdi, and A.~Belghith, ``Accurate 3d localization scheme based on
  active rfid tags for indoor environment,'' in \emph{2012 IEEE International
  Conference on RFID-Technologies and Applications (RFID-TA)}.\hskip 1em plus
  0.5em minus 0.4em\relax IEEE, 2012, pp. 378--382.

\bibitem{intro-uwb}
X.~Yubin, J.~Weilin, and S.~Xuejun, ``Toa estimate algorithm based uwb
  location,'' in \emph{2009 International Forum on Information Technology and
  Applications}, vol.~1.\hskip 1em plus 0.5em minus 0.4em\relax IEEE, 2009, pp.
  249--252.

\bibitem{toa}
J.~J. Leonard and H.~F. Durrant-Whyte, ``Simultaneous map building and
  localization for an autonomous mobile robot.'' in \emph{IROS}, vol.~3, 1991,
  pp. 1442--1447.

\bibitem{tdoa}
X.~Chen and Z.~Gao, ``Indoor ultrasonic positioning system of mobile robot
  based on tdoa ranging and improved trilateral algorithm,'' in \emph{2017 2nd
  International Conference on Image, Vision and Computing (ICIVC)}.\hskip 1em
  plus 0.5em minus 0.4em\relax IEEE, 2017, pp. 923--927.

\bibitem{rssi}
M.~Hamdollahzadeh, R.~Amiri, and F.~Behnia, ``Optimal sensor placement for
  multi-source aoa localisation with distance-dependent noise model,''
  \emph{IET Radar, Sonar \& Navigation}, vol.~13, no.~6, pp. 881--891, 2019.

\bibitem{challenge}
K.~Al~Nuaimi and H.~Kamel, ``A survey of indoor positioning systems and
  algorithms,'' in \emph{2011 international conference on innovations in
  information technology}.\hskip 1em plus 0.5em minus 0.4em\relax IEEE, 2011,
  pp. 185--190.

\bibitem{liu}
H.~Liu, H.~Darabi, P.~Banerjee, and J.~Liu, ``Survey of wireless indoor
  positioning techniques and systems,'' \emph{IEEE Transactions on Systems,
  Man, and Cybernetics, Part C (Applications and Reviews)}, vol.~37, no.~6, pp.
  1067--1080, 2007.

\bibitem{wireless-sensor}
I.~Amundson and X.~D. Koutsoukos, ``A survey on localization for mobile
  wireless sensor networks,'' in \emph{Mobile Entity Localization and Tracking
  in GPS-less Environnments: Second International Workshop, MELT 2009, Orlando,
  FL, USA, September 30, 2009. Proceedings}.\hskip 1em plus 0.5em minus
  0.4em\relax Springer, 2009, pp. 235--254.

\bibitem{smartphones}
P.~Davidson and R.~Pich{\'e}, ``A survey of selected indoor positioning methods
  for smartphones,'' \emph{IEEE Communications Surveys \& Tutorials}, vol.~19,
  no.~2, pp. 1347--1370, 2016.

\bibitem{emergency}
A.~F. G.~G. Ferreira, D.~M.~A. Fernandes, A.~P. Catarino, and J.~L. Monteiro,
  ``Localization and positioning systems for emergency responders: A survey,''
  \emph{IEEE Communications Surveys \& Tutorials}, vol.~19, no.~4, pp.
  2836--2870, 2017.

\bibitem{2019-survey}
F.~Zafari, A.~Gkelias, and K.~K. Leung, ``A survey of indoor localization
  systems and technologies,'' \emph{IEEE Communications Surveys \& Tutorials},
  vol.~21, no.~3, pp. 2568--2599, 2019.

\bibitem{2021-review}
H.~Obeidat, W.~Shuaieb, O.~Obeidat, and R.~Abd-Alhameed, ``A review of indoor
  localization techniques and wireless technologies,'' \emph{Wireless Personal
  Communications}, vol. 119, pp. 289--327, 2021.

\bibitem{trolley-first}
C.~Wang, X.~Mai, D.~Ho, T.~Liu, C.~Li, J.~Pan, and M.~Q.-H. Meng,
  ``Coarse-to-fine visual object catching strategy applied in autonomous
  airport baggage trolley collection,'' \emph{IEEE Sensors Journal}, vol.~21,
  no.~10, pp. 11\,844--11\,857, 2020.

\bibitem{real-time-plan}
J.~Wang and M.~Q.-H. Meng, ``Real-time decision making and path planning for
  robotic autonomous luggage trolley collection at airports,'' \emph{IEEE
  Transactions on Systems, Man, and Cybernetics: Systems}, vol.~52, no.~4, pp.
  2174--2183, 2022.

\bibitem{keypoints}
A.~Xiao, H.~Luan, Z.~Zhao, Y.~Hong, J.~Zhao, W.~Chen, J.~Wang, and M.~Q.-H.
  Meng, ``Robotic autonomous trolley collection with progressive perception and
  nonlinear model predictive control,'' in \emph{2022 International Conference
  on Robotics and Automation (ICRA)}, 2022, pp. 4480--4486.

\bibitem{epnp}
V.~Lepetit, F.~Moreno-Noguer, and P.~Fua, ``Ep n p: An accurate o (n) solution
  to the p n p problem,'' \emph{International journal of computer vision},
  vol.~81, pp. 155--166, 2009.

\bibitem{rfid-feature}
K.~Yamano, K.~Tanaka, M.~Hirayama, E.~Kondo, Y.~Kimuro, and M.~Matsumoto,
  ``Self-localization of mobile robots with rfid system by using support vector
  machine,'' in \emph{2004 IEEE/RSJ International Conference on Intelligent
  Robots and Systems (IROS)(IEEE Cat. No. 04CH37566)}, vol.~4.\hskip 1em plus
  0.5em minus 0.4em\relax IEEE, 2004, pp. 3756--3761.

\bibitem{rfid-tag}
R.~Tesoriero, J.~A. Gallud, M.~Lozano, and V.~M.~R. Penichet, ``Using active
  and passive rfid technology to support indoor location-aware systems,''
  \emph{IEEE Transactions on Consumer Electronics}, vol.~54, no.~2, pp.
  578--583, 2008.

\bibitem{manipulation}
A.~Collet, M.~Martinez, and S.~S. Srinivasa, ``The moped framework: Object
  recognition and pose estimation for manipulation,'' \emph{The international
  journal of robotics research}, vol.~30, no.~10, pp. 1284--1306, 2011.

\bibitem{grasping}
M.~Zhu, K.~G. Derpanis, Y.~Yang, S.~Brahmbhatt, M.~Zhang, C.~Phillips,
  M.~Lecce, and K.~Daniilidis, ``Single image 3d object detection and pose
  estimation for grasping,'' in \emph{2014 IEEE International Conference on
  Robotics and Automation (ICRA)}.\hskip 1em plus 0.5em minus 0.4em\relax IEEE,
  2014, pp. 3936--3943.

\bibitem{mobile-robot}
Z.~Cao, R.~Liu, C.~Yuen, A.~Athukorala, B.~K. Kiat~Ng, M.~Mathanraj, and U.-X.
  Tan, ``Relative localization of mobile robots with multiple ultra-wideband
  ranging measurements,'' in \emph{2021 IEEE/RSJ International Conference on
  Intelligent Robots and Systems (IROS)}, 2021, pp. 5857--5863.

\bibitem{multi-agent}
A.~Fishberg and J.~P. How, ``Multi-agent relative pose estimation with uwb and
  constrained communications,'' in \emph{2022 IEEE/RSJ International Conference
  on Intelligent Robots and Systems (IROS)}, 2022, pp. 778--785.

\bibitem{reflector}
J.~Yu, A.~Shen, X.~Luo, and T.~Xiao, ``Robot localization using laser
  positioning of reflectors based on icp,'' in \emph{2021 33rd Chinese Control
  and Decision Conference (CCDC)}, 2021, pp. 3145--3149.

\bibitem{kmeans}
S.~Na, L.~Xumin, and G.~Yong, ``Research on k-means clustering algorithm: An
  improved k-means clustering algorithm,'' in \emph{2010 Third International
  Symposium on Intelligent Information Technology and Security Informatics},
  2010, pp. 63--67.

\bibitem{cleaning}
J.~Li, Z.~Xu, D.~Zhu, K.~Dong, T.~Yan, Z.~Zeng, and S.~X. Yang, ``Bio-inspired
  intelligence with applications to robotics: a survey,'' \emph{Intelligence \&
  Robotics}, vol.~1, no.~1, pp. 58--83, 2021.

\end{thebibliography}

\end{document}